\documentclass[10pt, conference, compsocconf]{IEEEtran}
\IEEEoverridecommandlockouts
\usepackage{amsmath,amssymb,amsfonts}
\usepackage{algorithmic}
\usepackage{graphicx}
\usepackage{textcomp}
\usepackage{placeins}
\begin{document}

\title{Predicting the Computational Cost of Deep Learning Models
}

\author{\IEEEauthorblockN{Daniel Justus}
\IEEEauthorblockA{
\textit{Digital Catapult}\\
London, United Kingdom \\
daniel.justus@digicatapult.org.uk}
\and
\IEEEauthorblockN{John Brennan}
\IEEEauthorblockA{\textit{School of Computing} \\
\textit{Newcastle University}\\
Newcastle, UK \\
john.brennan@ncl.ac.uk}
\and
\IEEEauthorblockN{Stephen Bonner}
\IEEEauthorblockA{\textit{Department of Computer Science} \\
\textit{Durham University}\\
Durham, UK \\
s.a.r.bonner@durham.ac.uk}
\and
\IEEEauthorblockN{Andrew Stephen McGough}
\IEEEauthorblockA{\textit{School of Computing} \\
\textit{Newcastle University}\\
Newcastle, UK \\
stephen.mcgough@newcastle.ac.uk}
}

\maketitle

\begin{abstract}
Deep learning is rapidly becoming a go-to tool for many artificial intelligence problems due to its ability to outperform other approaches and even humans at many problems. Despite its popularity we are still unable to accurately predict the time it will take to train a deep learning network to solve a given problem. This training time can be seen as the product of the training time per epoch and the number of epochs which need to be performed to reach the desired level of accuracy. Some work has been carried out to predict the training time for an epoch -- most have been based around the assumption that the training time is linearly related to the number of floating point operations required. However, this relationship is not true and becomes exacerbated in cases where other activities start to dominate the execution time. Such as the time to load data from memory or loss of performance due to non-optimal parallel execution. In this work we propose an alternative approach in which we train a deep learning network to predict the execution time for parts of a deep learning network. Timings for these individual parts can then be combined to provide a prediction for the whole execution time. This has advantages over linear approaches as it can model more complex scenarios. But, also, it has the ability to predict execution times for scenarios unseen in the training data. Therefore, our approach can be used not only to infer the execution time for a batch, or entire epoch, but it can also support making a well-informed choice for the appropriate hardware and  model.
\end{abstract}


\section{Introduction}\label{intro}
Deep learning has flourished over recent years, in no small part due to its super-human ability to learn patterns within data. This has been demonstrated in tasks such as image recognition~\cite{hu2017} through to beating the world champion at Go~\cite{silver}. This ability to outperform humans has its price -- that of the computational cost\footnote{Referred to here as execution time without loss of generality.} of training these complex networks and requiring significant volumes of data to reach the level of accuracy required to `beat' humans or just provide the required level of accuracy. The situation is compounded by many factors: identifying an `optimal' network architecture which has the chance to perform as desired, determining the best set of hyper-parameters for the network, determining the volume of data required for training along with determining {\em a priori} if the training can be performed within the required cost\footnote{Here we take a broad view of cost including financial and execution cost.} envelope.

The prediction of execution time is significant as it will not only allow us to predict the cost of performing the training but also the number of scenarios and hyperparameter settings that can be tested and thus eventually the accuracy of a model itself. Therefore, it is desirable to understand what the execution time and the associated costs will be {\em a priori} to training in order to determine value for money or to alter one's strategy to reduce execution time and hence cost.

The prediction of execution time can be decomposed into two main elements: the execution time of a single epoch -- a single forwards and backwards pass through all of the training data -- and predicting the number of epochs which will be required in order to reach a desired level of accuracy. Both are worthy of research. However, in this work we address the former case of estimating epoch execution time. In the future we will address the complementary secondary case. 

Although our work here focuses on the training times for deep learning networks our approach could be as easily applied to predicting the execution time for using a trained deep learning network to infer predictions by just looking at the feed-forward phase of the training.

Prior work in the area of epoch execution time prediction has focused on Big {\em O()} notation -- primarily based on the number of floating point operations performed within an epoch. For example the PALEO~\cite{Qi2016} system computes the number of floating point operations required for an epoch and multiplies this by a scaling factor derived from testing the floating point operation speed on a given system. However, this fails to take into account numerous other operations that are performed which do not scale linearly with the number of floating point operations. Nor does it take into account various system limitations, for example, less than optimal performance of the GPU due to data size or ability to make full use of all GPU cores simultaneously.

By contrast, here we propose training a deep neural network on `features' derived from the computational resource used, the network being trained and the data used for training. By performing such a process we can `learn' a representation of the training execution time which is more complex than the linear models which have been used thus far. Further, given enough training data from a wide-range of exemplar computational resources, neural networks and input data sets, it will, we believe, be possible to predict, to a reasonable level of accuracy, the training execution time for a problem case not previously seen by our approach -- be this different hardware, deep learning network, data or combination of these.

\begin{figure}[btp]
	\centerline{\includegraphics[width=0.47\textwidth]{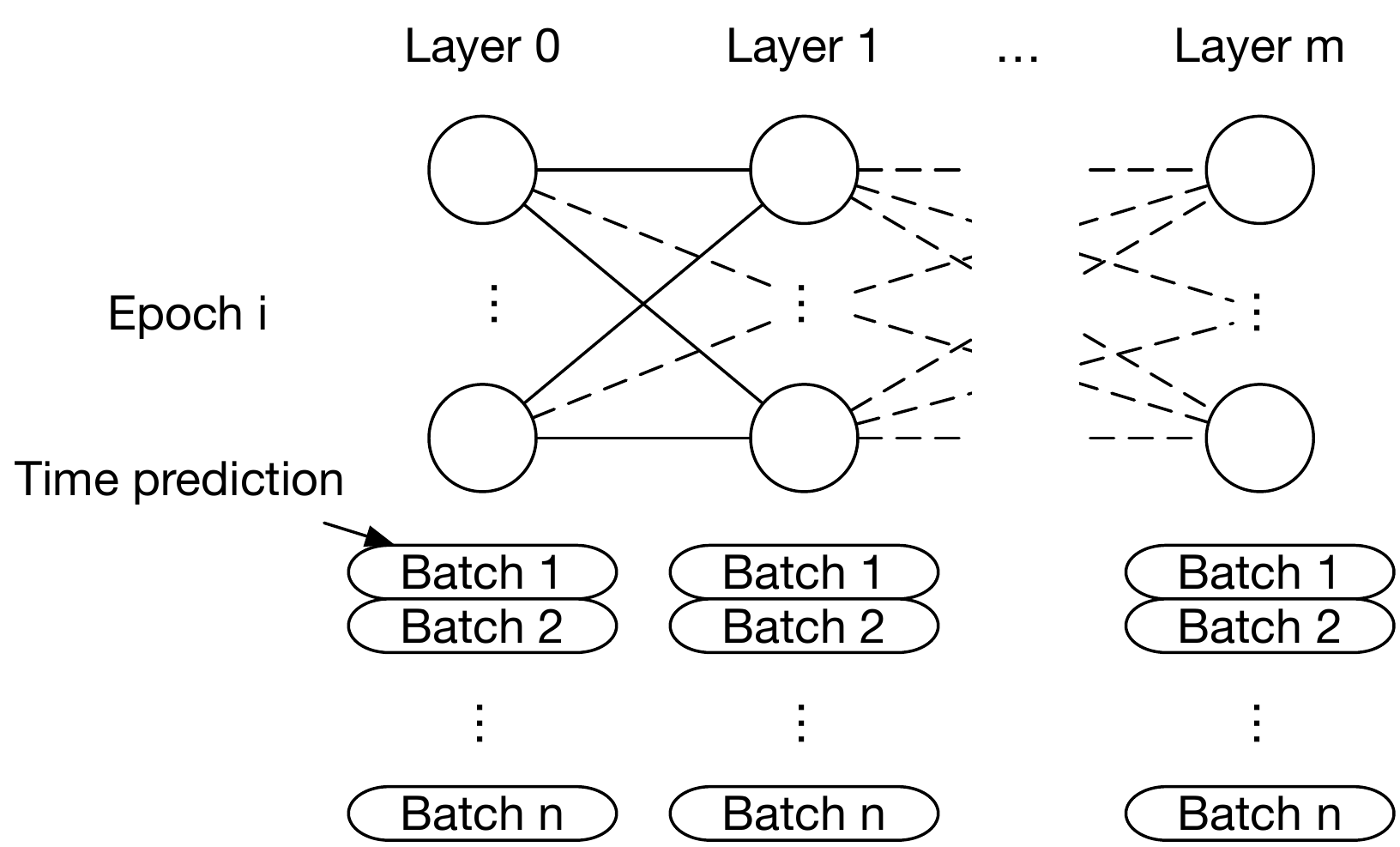}}
	\caption{Overview for generating a timing prediction for a full epoch}
	\label{fig:epoch_batch}
\end{figure}

For our approach we build a generally applicable, data driven model for predicting the execution time for commonly used layers in deep neural networks. From this we deduce the execution time required for one training step (forward and backward pass) for processing an individual batch of data. The overall execution time can then be calculated as the time per batch multiplied by the number of batches -- see Figure~\ref{fig:epoch_batch}. In this figure each layer of the neural network is displayed left to right and the figure concentrates on just one epoch. Data is fed through the network in batches and it is the timing of one single batch that we are predicting here. It should be noted that although the figure here represents a fully connected Multi-Layer Perceptron network we are not restricted to such networks. Our approach works equally well for other network layer types, such as convolutions or pooling layers.

As convolutions and fully connected layers (vector-matrix multiplications) make up the most significant part of the execution time in the majority of deep neural networks, we focus on these types of operations. However, we are confident that the same process will be equally applicable to other layer types. Further, as GPU cards are commonly used for performing deep learning we focus this paper on the performance of these cards. However, our approach would be equally applicable to CPU, TPU or even IPU deep learning.

We see our contribution in this work to be:
\begin{itemize}
	\item The categorisation of features which may influence the performance of deep learning training execution times.
	\item Predicting the execution time for a deep learning network through the use of deep learning.
	\item Comparison of performance of predicted execution times for different hardware.
\end{itemize}

The rest of this paper is structured as follows. In Section~\ref{related} we discuss the related work to our approach. We present a motivation for moving beyond a linear model in Section~\ref{motivation}. The different features which can be captured from a deep learning network and the environment in which it runs are discussed in Section~\ref{training}. Section~\ref{method} discusses the methodology used in this work including how we combine individual batches to compute the whole network execution time. We define the experiments performed and the features used in Section~\ref{experiment}. The results of our approach are presented and discussed in Section~\ref{results}. Finally, we evaluate our approach and provide conclusions in Section~\ref{conc}.

\section{Related Work}
\label{related}
The complexity of machine learning models, and in particular of deep learning models has been steadily increasing over recent years. This is in no small part a consequence of the increasing number of layers which are used within the deep learning network -- for example the the winner of ImageNet in 2012 (AlexNet~\cite{Krizhevsky2012}) contained just 8 layers, whilst by 2015 this had increased to 152 layers (ResNet~\cite{He2015}). This is being compounded by the fact that we are now throwing much more data through these deep learning networks~\cite{Zhang2018}.

With this rapid increase in complexity, along with the volumes of data we wish to process, there is a corresponding increase in the execution time required to train a network. Irrespective of whether the network training is performed in the cloud or on locally provisioned resources this will lead to excessive training costs. Up-to-date GPUs provisioned in the cloud for deep learning often cost several dollars per hour and the expenditure for training a deep neural network on GPUs can easily reach the order of thousands of dollars. Similarly, the acquisition of locally provisioned resources plus expenses for electricity and air conditioning can cause significant costs.

 Despite there being a significant financial cost involved with training a deep learning network, there has been little research into predicting the execution time and the choice of optimal hardware. Benchmarking initiatives like DAWNbench~\cite{Coleman2017} and MLPerf\footnote{https://mlperf.org/} aim at quantifying performance of different hardware chipsets when training a number of machine learning model architectures. However, by design these approaches are limited to a few reference architectures. Work has been done which shows that one can predict the execution time for a new model by attempting to compare the model to similar models which have known performance~\cite{Adolf2016}, however, this can only yield very coarse estimates.

A different approach is to generate a performance prediction from timings generated from the individual floating point operations that are executed during a training step \cite{Qi2016}. This is justified by the fact that most of the deep learning approach is based around linear algebra operations using floating point mathematics, where the number of floating point operations performed can be easily computed. However, due to the lack of perfect parallelism of computations on GPUs, the fact that non-floating point operations are used and the data transfer times between the GPU and main memory, the execution time only scales approximately linearly with the number of floating point operations performed. Qi {\em et al.}~\cite{Qi2016} attempt to compensate for this through a scaling derived from observing real deep learning training, however, this still assumes an even distribution of floating point and non-floating point work across all deep learning.
%

\section{Motivation}
\label{motivation}
The execution time required during a forward pass through a neural network is bounded from below by the number of floating point operations (FLOPs)~\cite{jouppi2017}. This FLOP count depends on the deep neural network architecture and the amount of data. The time required for each of these FLOPs depends on the hardware specifications. Similarly, communication times have a generally linear relationship with data size, where data transfers are properly managed and memory bandwidth is properly utilised.

This type of linear model is where efforts have been focused in recent literature~\cite{Qi2016}. However, other features such as activation functions and the optimizer used introduce sources of nonlinearities into the system. Moreover, a full utilisation of the available compute resources and memory bandwidth cannot be achieved in all situations. Therefore, even more sophisticated approaches yield unsatisfactory results and generalise badly for different kinds of hardware.

Our work aims to fill this gap by providing a prediction framework that is capable of dealing with these non-linearities, providing accurate results while also generalising to previously unseen neural network models or hardware.

\section{Training Features}
\label{training}

We define here the features which could influence the prediction of execution times when performing training. We categorise these features into layer features, layer specific features, implementation features and hardware features. Each of these categories can contain an almost endless list of features. As such we define here a core subset of those features but argue that other features could easily be added. A full analysis of all available features and the impact they have on the accuracy of prediction is beyond the scope of this work.

\subsection{Layer Features}
These relate to those features of a particular layer within the neural network and in particular to the hyperparameters related to that layer. These include, but are not limited to:
\begin{itemize}
	\item {\bf Activation function} used on the individual neurone. These can include none, ReLU, softmax, sigmoid and tanh. They can be encoded into the feature set using one-hot encoding.
	\item {\bf Optimiser} used for locating a minima within the loss function space. These can include Gradient Descent, Adadelta, Adagrad, Momentum, Adam and RMS Prop. They can be encoded into the feature set using one-hot encoding.
	\item {\bf Batch size} representing the number of training samples which are processed together as part of the same batch.
\end{itemize}

It should be noted that each individual layer within the network may possess different values for these features. As each layer is predicted independently this is not a problem.

\subsection{Layer Specific Features}
Here we discuss the features which are unique to a particular type of layer within the neural network.

\subsubsection{MLP features} as this represents a fully connected layer within the network we are concerned here with the number of neurones not only within this layer but the preceding and following layers too.
\begin{itemize}
	\item {\bf Number of inputs} to the layer. As all layers are fully connected this value is effectively the number of outputs from the previous layer. 
	\item {\bf Number of neurones} within this layer. Which is equivalent to the number of outputs of the layer.
\end{itemize}

\subsubsection{Convolutional features} relate to those features relevant for convolutional layers within a network. These include:
\begin{itemize}
	\item {\bf Matrix size} representing the size of the input data to be trained on.
	\item {\bf Kernel size} representing the size of the filter applied to the image data.
	\item {\bf Input depth} is the number of channels or layers in the input data.
	\item {\bf Output depth} is the number of channels or layers in the output data. 
	\item {\bf Stride size} the size of the stride to be made with the convolution kernel.
	\item {\bf Input padding} the number of border layers of zeros added to the outside of the matrix in order to allow proper analysis of edge pixels.
\end{itemize}

\subsubsection{Pooling features} are those features unique to the pooling layer within a network. 

The features include: 
\begin{itemize}
	\item {\bf Kernel size} the size of the pooling kernel.	
	\item {\bf Stride size} the size of the stride to be made with the pooling kernel.
	\item {\bf Input padding} the number of border layers of zeros added to the outside of the matrix in order to allow proper analysis of edge pixels.
\end{itemize}

\subsubsection{Recurrent features} representing those features which define a RNN. A RNN will contain the same features outlined for MLP -- i.e. number of inputs and number of neurones. In addition they also include:
\begin{itemize}
	\item {\bf Recurrence type} as there are many types of recurrence we define a one-hot encoding of these types, including: default, LSTM, and GRU.
	\item {\bf Bidirectional} a binary value indicating if the RNN is bidirectional or not.
\end{itemize}

Additional feature sets and individual features can be added to our model, however, these are considered to be the core features covering the majority of deep neural network in use today.

%

\subsection{Hardware Features} 
Hardware features are those describing both the GPU card(s) used and the system these cards are housed within:
\begin{itemize}
	\item {\bf GPU technology} identifying the chip manufacturer and the chip technology used. This can include the  NVIDIA microarchitectures Turing, Volta, Pascal, Maxwell, and Kepler, but also other manufacturers. This can be encoded as a one-hot encoding. It should be noted that this only describes the GPU technology not the individual cards. Thus allowing for cards from the same generation to be grouped together.
	\item {\bf GPU count} the number of GPUs in the system. If there are multiple GPUs per card then this is the number of GPUs.
	\item {\bf GPU memory} is the memory available per GPU card. In the case of multiple GPUs built into a single card this is the memory per GPU.
	\item {\bf GPU clock speed} recorded in Hz.
	\item {\bf GPU memory bandwidth} recorded in GB/s.
	\item {\bf GPU core count} the number of processing units. In the case of NVIDIA GPUs this is the number of cuda cores.
	\item {\bf GPU peak performance} recorded in GFLOPS, a result of GPU clock speed and GPU core count.
	\item {\bf Card connectivity} this is a one-hot encoding of the different interconnects available for the GPU. This can include: PCIe3.0 x16, PCIe3.0 x4, NVLink.
\end{itemize}

Although the focus here is on GPU based deep learning this set of features could easily be expanded to cater for other hardware types.

\subsection{Training Space}
It should be noted that as this feature space contains an extremely large number of possible combinations it is not feasible to train a network on all values. As such we train our networks through a random subsample of the feature space. 


\section{Methodology}
\label{method}
\subsection{General Considerations}

Our approach here is to break up each deep learning network into single components, considering individual layers as the atomic operations we are going to use for performance prediction (see Figure~\ref{fig:epoch_batch}). We construct a random selection of these atomic operations embedded within the simplest network we can produce. Each of these atomic operations is then executed either using forward or forwards and backwards passes and the execution times are recorded from multiple executions. The feature set for the atomic operation along with the execution timings are then used to train a fully connected feed forward network. This network can then be used for predicting the execution time for a new operation based just on the feature set.

Once a prediction has been made for an individual operation these can be combined together and across layers in order to provide a prediction for the overall performance of the deep learning network. By working with these atomic operations we help to reduce the computational time to train our approach whilst also maximising the range of layer types that we can predict. Moreover, to add a new layer type is a simple case of benchmarking batches of that layer type and the re-training a deep learning model with this extra feature and performance information added. As the deep learning predictor network is relatively simple the re-training time is not seen as being significant.

\subsection{Model Architecture and Training Procedure}\label{model}
In order to obtain a prediction model for the execution time for training a deep learning network we have developed our own fully connected feed forward deep learning network which is trained on the feature sets defined above and results from actual training runs of deep learning networks. Our neural network architecture consists of $m$ layers with $j_n$ neutrons in layer $n$. Each of these layers is followed by a dropout layer, only used in training, with the final dense layer producing a single output.

In order to produce an accurate deep learning predictor we evaluated numerous runs of our predictor using various parameters. The parameters considered during this optimization process were; $m$, $j_n$ ($n \in [0,m]$), dropout rate, loss function and the model optimizer.

\subsection{Full Model Predictions}
To predict the computation time for a single epoch of a deep network we can first compute the time that is required for a forward and backward pass on a single batch:
$$
T_b = \sum_{i=0}^{l} b_{M(i)}
$$
where $l$ is the number of layers in the deep neural network and $b_{M(i)}$ is the batch execution time estimate, generated by our prediction approach, for layer $i$, where $M(i)$ is of type of layer $i$. It should be noted here that $b_{M(i)}$ should also be parameterised by the other features we use when training our network, however, for simplicity we have not listed those here. Then to compute the total execution time for a single epoch of the deep learning network we can compute:
$$
E = p T_b
$$
where $p$ is the number of batches required to process the data.

To compute the total time for training a network would also require a prediction for the number of epochs required -- which is beyond the scope of this work. However, as many deep learning users currently use a fixed number of epochs (for example 100) an estimate of this form can easily be made.

\begin{table*}[htb]
	\center\vskip 8pt
	\begin{tabular}{l|c|l|l|l|l|l}
		{\bf Name} & {\bf Provisioning} & {\bf Cuda cores} & {\bf Clock (boost)} & {\bf Memory} & {\bf GPU memory bandwidth} & {\bf Bus}\\
		\hline
		V100 & Local & 5120 & 1455 MHz & 16 GB HBM2 & 900 GB/s & NVLink\\
		P100 & Cloud & 3584 & 1303 MHz & 16 GB HBM2 & 732 GB/s & PCIe\\
		GTX1080Ti & Local & 3584 & 1582 MHz & 11 GB GDDR5X & 484 GB/s & PCIe\\
		M60 & Cloud & 4096 & 1178 MHz & 16 GB GDDR5 & 320 GB/s &  PCIe\\
		K80 & Cloud & 2496 & 875 MHz & 12 GB GDDR5 & 240 GB/s & PCIe\\
		K40 & Local & 2880 & 875 MHz & 12 GB GDDR5 & 288 GB/s & PCIe\\
	\end{tabular}
	\vskip 4pt
	\caption{Specification of hardware tested}
	\label{tab:hw}
\end{table*}

\subsection{Comparison Metrics}
In order to evaluate our approach we compare for a given (unseen) test data set the predicted execution times along with the actual execution times. If our approach is efficient we will see a strong correlation between these two data sets. In addition to comparing actual values we also look at the root mean squared error (RMSE) between the predicted and actual values, where smaller values indicate better predictions.

\section{Experimental Setup}
\label{experiment}
We describe here the layers, and hardware we used for the evaluation of our approach. As the number of hardware platforms available for testing was relatively small we train models both on individual GPU cards as well as on a combined set of GPU cards and use this to compare between the different approaches. We anticipate that with the inclusion of more platforms a single model would not only be as accurate, but would be able to predict for previously unseen hardware. However, as our available hardware is relatively limited we anticipate that at present the combined model may not achieve a favourable accuracy when compared to individual models.

\subsection{Hardware}
Table \ref{tab:hw} outlines the different hardware used for experimentation. We shall use the name when referring to the hardware in the following discussion; provisioning indicates if the hardware was locally owned or provisioned from a Cloud provider, whilst specification indicates the main characteristics of the system tested.

\subsection{Test Case: Fully Connected Layer}
In order to evaluate our deep learning predictor for fully connected layers we used the Tensorflow implementation \mbox{\textit{tensorflow.layers.dense}} to generate fully connected layers. We tested 25,000 of the possible parameter combinations from a search space of 30,064,771,072 possible combinations. These were derived from batch sizes in the range 1 to 64, input and output dimensions in the range  1 to 4096.

\subsection{Test case: Convolutional Neural Layer}
To assess convolutional layers we tested the \mbox{\textit{tensorflow.layers.conv2d}} operation with different batch sizes in the range of 1 to 64, matrix sizes in the range of 1x1 to 512x512, kernel sizes in the range of 1x1 to 7x7, and up to \(\frac{10000}{\text{matrix size}}\) input and output layers. Moreover, we tested the convolutional layers under the condition of different stride sizes, input paddings and with or without adding a bias. Out of 10,038,745,006,080 possible combinations of these parameters we benchmarked up to 50,000 randomly chosen combinations and used these results to train our deep neural network to then predict arbitrary parameter settings.

\subsection{Data Collection and Preparation}
Feature sets were selected at random using a uniform distribution approach with no interdependence on previously selected features. For each feature set we performed 5 benchmark runs and used the median result as input to our model. The collected data was split into a training data set (80\%), a test data set (10\%) and a validation data set (10\%). 

\subsection{Deep Learning Prediction Network}
We performed a parameter search of different fully connected feed forward deep learning networks and identified different required depths for different problems. Likewise, having one dropout layer after the final hidden layer along with using the ReLU~\cite{relu} activation function and using L2 regularisation~\cite{l2r} with a regularisation constant of $10^{-5}$, used in every layer of the network, helped to maximise the generalisability and the overall accuracy. To emphasize the importance of accurate results on frequently occurring computationally cheap operations we used a root mean square logarithmic error in the loss function. The Adam~\cite{adam} optimiser with a decaying learning rate was used to minimise the loss.

The model was trained for 300 epochs with a batch size of 128 for each of the benchmark runs. The initial learning rate of 0.1 was reduced every 40 epochs by a factor of 2. The loss on the test set was calculated after each epoch of training on the full test data set. All hyperparameter tuning was performed based on the loss returned from the test data. 

\subsection{Implementation Details}
To allow comparability throughout all tested systems we containerised our code using Docker\footnote{All resources are available on https://github.com/CDECatapult/ml-performance-prediction.git}. We aimed at developing a model that is capable of predicting a wide range of common operations used in the Tensorflow machine learning framework on different GPUs. We used the containerised version of tensorflow1.10.1-gpu as provided on Dockerhub\footnote{https://hub.docker.com/r/tensorflow/tensorflow/} for all our experiments.

For all experiments we generated training data from the model to be predicted using a randomly selected activation function from a choice of: \textit{none}, \textit{tensorflow.nn.relu}, \textit{tensorflow.nn.sigmoid} and \textit{tensorflow.nn.tanh}. We also randomly selected the optimizers from: \textit{tensorflow.train.$\{$GradientDescentOptimizer, AdadeltaOptimizer, AdagradOptimizer, MomentumOptimizer, AdamOptimizer, RMSPropOptimizer$\}$}. 50\% of all experiments were performed as pure forward passes without any backpropagation.

\section{Results}
\label{results}
We present here results for training models using the features we have defined previously within this work. We first demonstrate the capability of our approach to predict the performance of a GPU when trained on data only from that GPU and compare our approach against a linear regressor. We then test our deep learning predictor when it has been trained on data from multiple GPU types. In this second case we use the trained model to predict execution times for a GPU type not seen as part of the training process. In the last case we evaluate how good our approach is by predicting the entire training execution time for a single batch across all layers of a real deep learning network. In all experiments dropouts were not used unless otherwise specified.

\subsection{Estimating the training time for a fully connected or convolutional layer}
Here we train a deep learning predictor for each GPU type separately from each other. This allows us to validate if our approach will allow us to predict the execution time without taking the features of hardware into account. Thus, this is both a simpler model and removes the issue that we had relatively few cards on which to train. It was anticipated that these predictions would be more accurate than those generated for a combined model of multiple GPU types.

\begin{figure}[b]
    \vskip -12pt
	\centerline{\includegraphics[width=8cm]{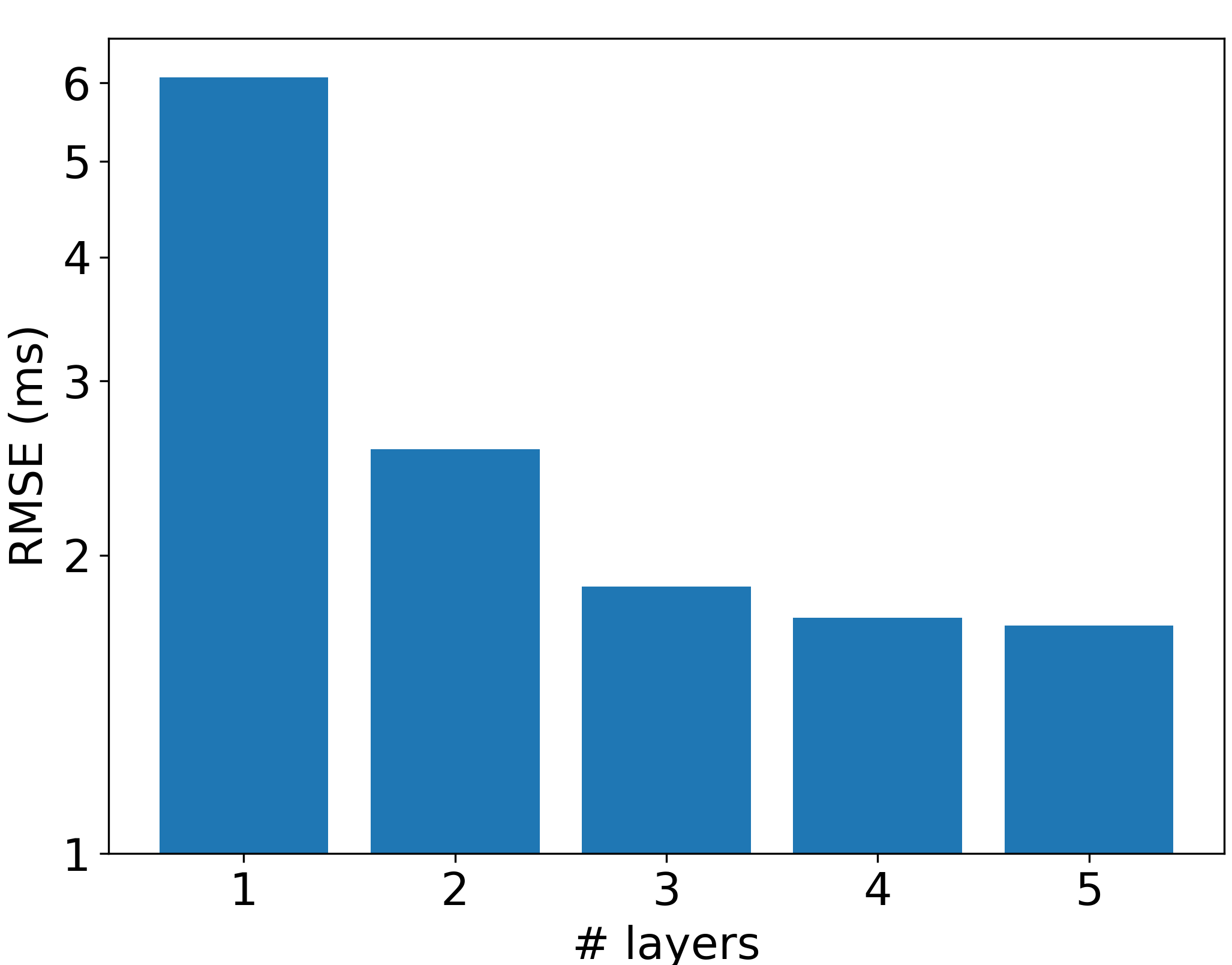}}
	\caption{RMSE of execution time predictions for convolutional layers on an V100 GPU using models with different numbers of hidden layers}
	\label{fig:V100_size-vs-error}
\end{figure}

We first assess the minimal complexity of a prediction model that is required to predict the execution times for convolutional layers. Figure~\ref{fig:V100_size-vs-error} depicts the RMSE when predicting performance of an NVIDIA Tesla V100 GPU using models with different numbers of hidden layers. Although five layers has the lowest RMSE the benefit over a four layer network is negligible in comparison to the increase in training time. Following these results we use neural networks with four hidden layers to predict the performance of individual GPUs.

\begin{figure*}[hbt]
	\centerline{\includegraphics[width=1.0\textwidth]{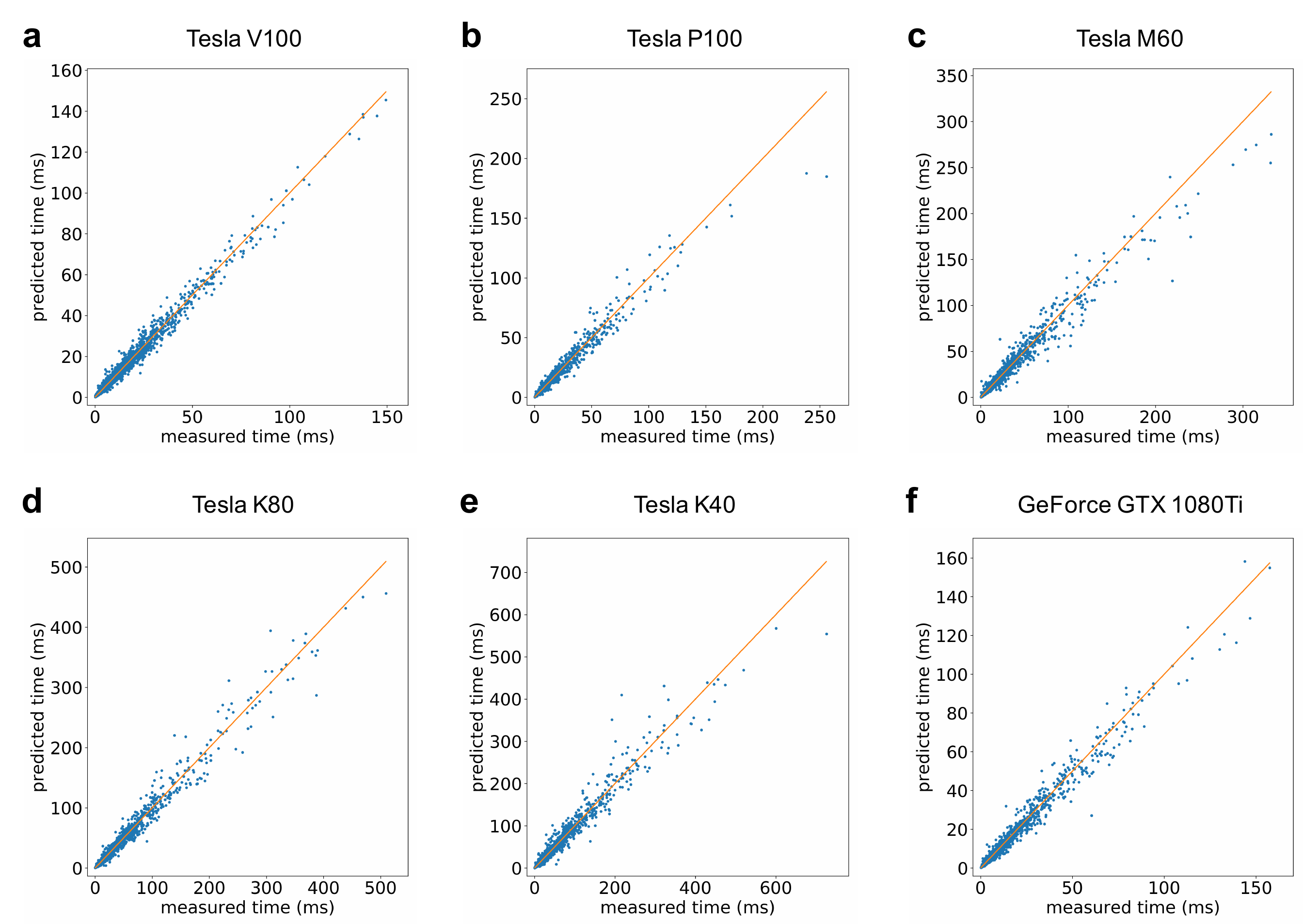}}
	\caption{Predicted time for convolutions versus measured time. \textbf{a)} Tesla V100, RMSE of prediction 1.73 ms. \textbf{b)} Tesla P100, RMSE 3.32 ms. \textbf{c)} Tesla M60, RMSE 6.07 ms. \textbf{d)} Tesla K80, RMSE 7.84 ms. \textbf{e)} Tesla K40, RMSE 11.90 ms. \textbf{f)} Geforce GTX1080Ti, RMSE 2.55 ms}
	\label{fig:prediction_GPUs_conv}
\end{figure*}

Figure~\ref{fig:prediction_GPUs_conv} depicts the comparison between predicted and actual execution training times for four different GPU cards for convolutional layers. Namely a) NVIDIA Tesla V100, b) NVIDIA Tesla P100, c) NVIDIA Tesla M60, d) NVIDIA Tesla K80, e) NVIDIA Tesla K40, and f) NVIDIA Geforce GTX1080Ti. These result in a RMSE in each case, respectively, of a) 1.73 ms, b) 3.32 ms, c) 6.07 ms, d) 7.84 ms, e) 11.90 ms, and f) 2.55 ms. It should be noted that older GPUs show longer execution times, and accordingly larger absolute deviations in the predictions.

The execution times for the fully connected layers were predicted similarly. Hence, we do not present the graphs here. We do report the RMSE as a) 0.048 ms, b) 0.033 ms, c) 0.031 d) 0.145 ms, e) 0.167 ms, and f) 0.034 ms respectively. These relatively small errors can be explained by smaller average execution times for the fully connected layers tested.

\subsection{Comparison with other prediction approaches}
\begin{figure*}[tb]
	\centerline{\includegraphics[width=\textwidth]{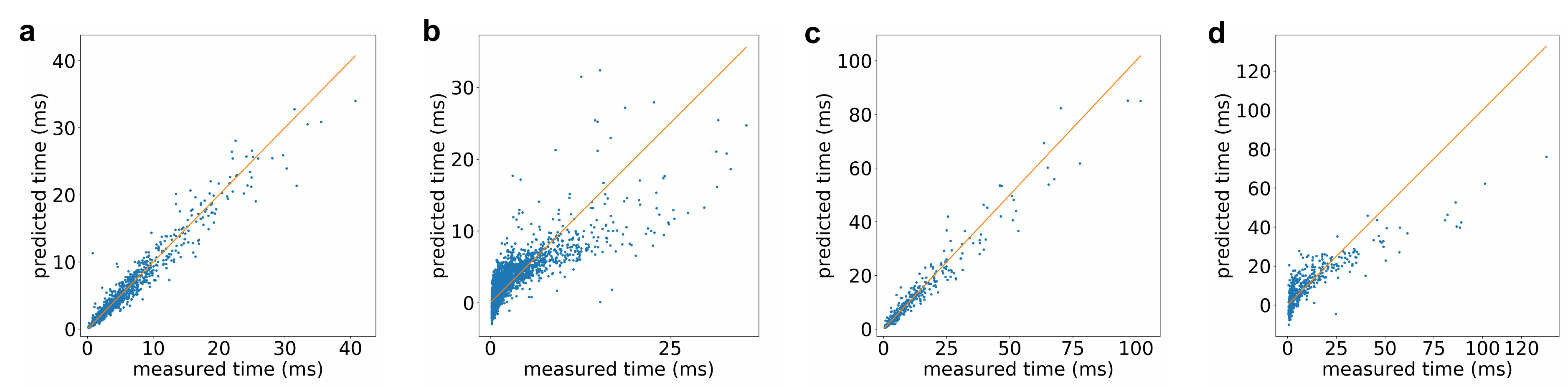}}
	\caption{Predicted execution time for convolutions on a Tesla V100 versus the measured time. \textbf{a)} Deep neural network model for a forward pass only, RMSE 0.83 ms. \textbf{b)} Linear regression model for a forward pass only, RMSE 2.42 ms. \textbf{c)} Deep neural network model for a forward and backward pass with stochastic gradient descent, RMSE 3.18 ms. \textbf{d)} Linear regression model for a forward and backward pass with stochastic gradient descent, RMSE 8.95 ms.}
	\label{fig:DNNvsLR}
\end{figure*}

We compare our approach here with a simple linear regression model~\cite{linear}. As the linear regression model is not able to handle our one-hot encoded features we cannot use them here. Thus, we separately evaluate predictions for only forward passes (Figure~\ref{fig:DNNvsLR}a,b), and predictions for forward and backward passes with stochastic gradient descent~\cite{kiefer1952} (Figure~\ref{fig:DNNvsLR}c,d). In both cases we use the number of required floating point operations as additional input feature to the linear model to mitigate the non-linearities. We assess the accuracy of the models by comparing the predicted execution time with the actual execution time for different convolutional layers. 

Figure~\ref{fig:DNNvsLR}a depicts the results for using our deep learning predictor to predict the execution time for a feed-forward pass, whilst Figure~\ref{fig:DNNvsLR}b is for the linear regressor. The linear regressor clearly fails to capture the complexity of the model and produces a large inaccuracy (RMSE 2.42 ms). By contrast the deep learning predictor produces a fairly consistent prediction in comparison to the actual training time (RMSE 0.83 ms). This indicates that a deep learning based predictor is much better to use for such cases. This is likely to be a consequence of the linear regressor being unable to account for non-linear relationships between features and execution time.

Figure~\ref{fig:DNNvsLR}c demonstrates that our deep learning predictor consistently generates good estimates for the execution times for forward and backward passes across the range of actual execution times with a root mean squared error of 3.18 ms. However, Figure~\ref{fig:DNNvsLR}d shows that the linear regression method becomes progressively worse as the actual execution time increases, leading to a RMSE  of 8.95 ms. Again, the linear regression approach is modelling for the majority of observed data points and missing the non-linear effects of longer run-times.

While additional feature engineering that goes beyond the number of floating point operations might overcome some of the problems with the linear model, it cannot be expected to capture the full complexity of the execution time prediction. Moreover, by using just forward passes or a single optimiser may be beneficial for a linear approach, our deep learning approach can also handle different optimisers simultaneously.

\subsection{A general model constructed from multiple cards}\label{general_model}
\begin{figure}[hbt]
	\centerline{\includegraphics[width=8cm]{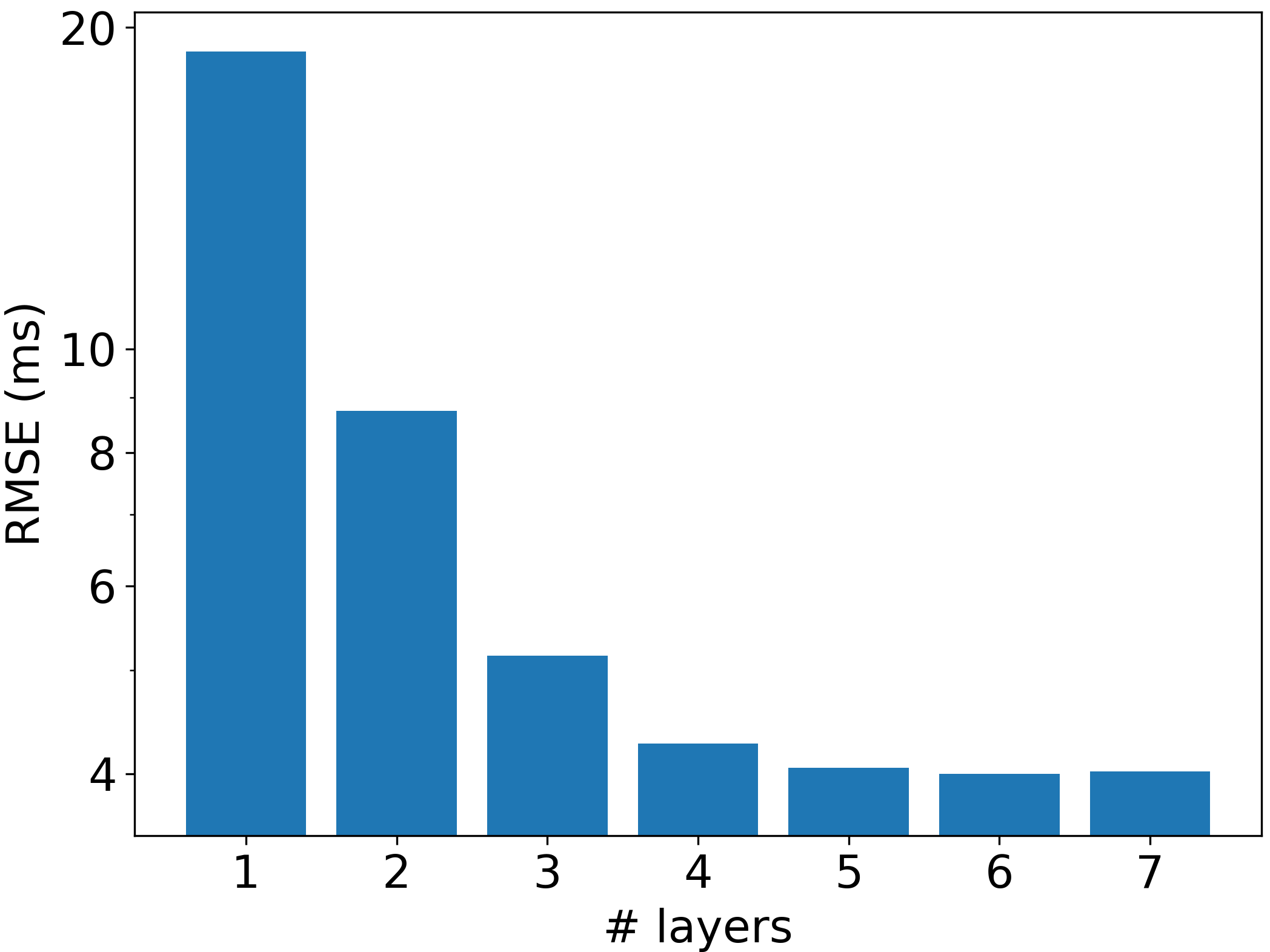}}
	\caption{RMSE of execution time predictions for convolutional layers on an NVIDIA Tesla V100 GPU using models with different numbers of hidden layers}
	\label{fig:all_size-vs-error}
\end{figure}

We now evaluate the results for our model trained on data from multiple GPUs. The features of the model were expanded to include GPU memory bandwidth, GPU clock frequency and the number of CUDA cores. Since these additional features might require a more complex model we reassess the required number of hidden layers for this case. The results shown in Figure~\ref{fig:all_size-vs-error} suggest that using 6 hidden layers yield the best results.

\begin{figure}[htb]
	\centerline{\includegraphics[width=8cm]{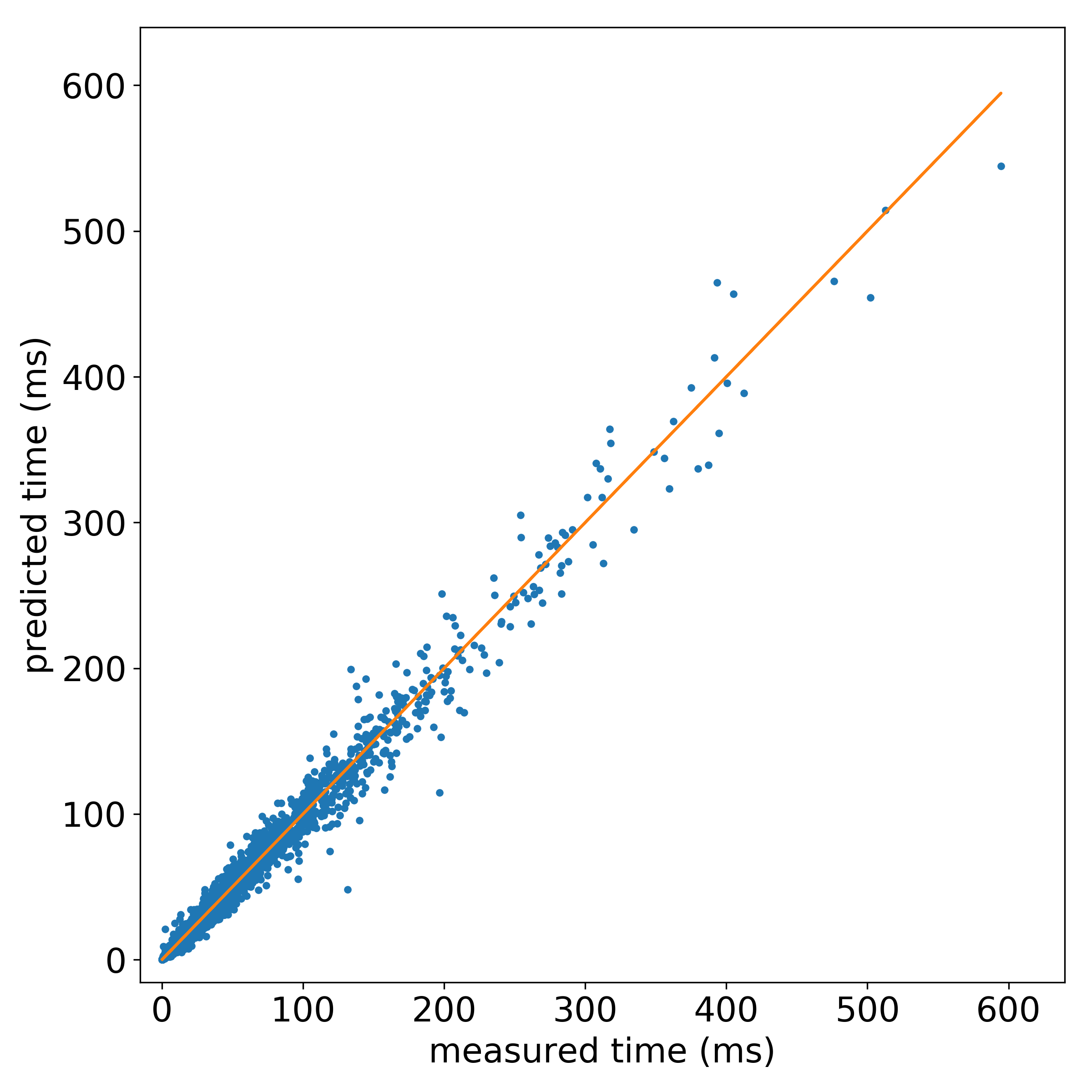}}
	\caption{Predicted time for convolutions versus measured time for a model that has been trained on data from all available GPUs. The RMSE is 3.88 ms.}
	\label{fig:Model_all}
\end{figure}

We first trained our model on data from all available GPUs to test the viability of this approach. Since the limited memory of some of the GPUs did not allow for using larger batch sizes, in this experiment we only used data with batch sizes of up to 32. Figure~\ref{fig:Model_all} depicts the comparison between predicted and actual execution training times for this setting. The quality of the predictions with a RMSE of 3.88 ms shows that this model is capable of inferring executions times of convolutions on different GPUs.

The more interesting case, however, is predicting of the performance of different, unseen hardware, using just the above features. To test whether our approach can be used for this general performance prediction, we trained a model on the data from just five of the GPUs. The data for the sixth GPU was held back to provide an unseen hardware case.

\begin{figure*}[htbp]
	\centerline{\includegraphics[width=1.0\textwidth]{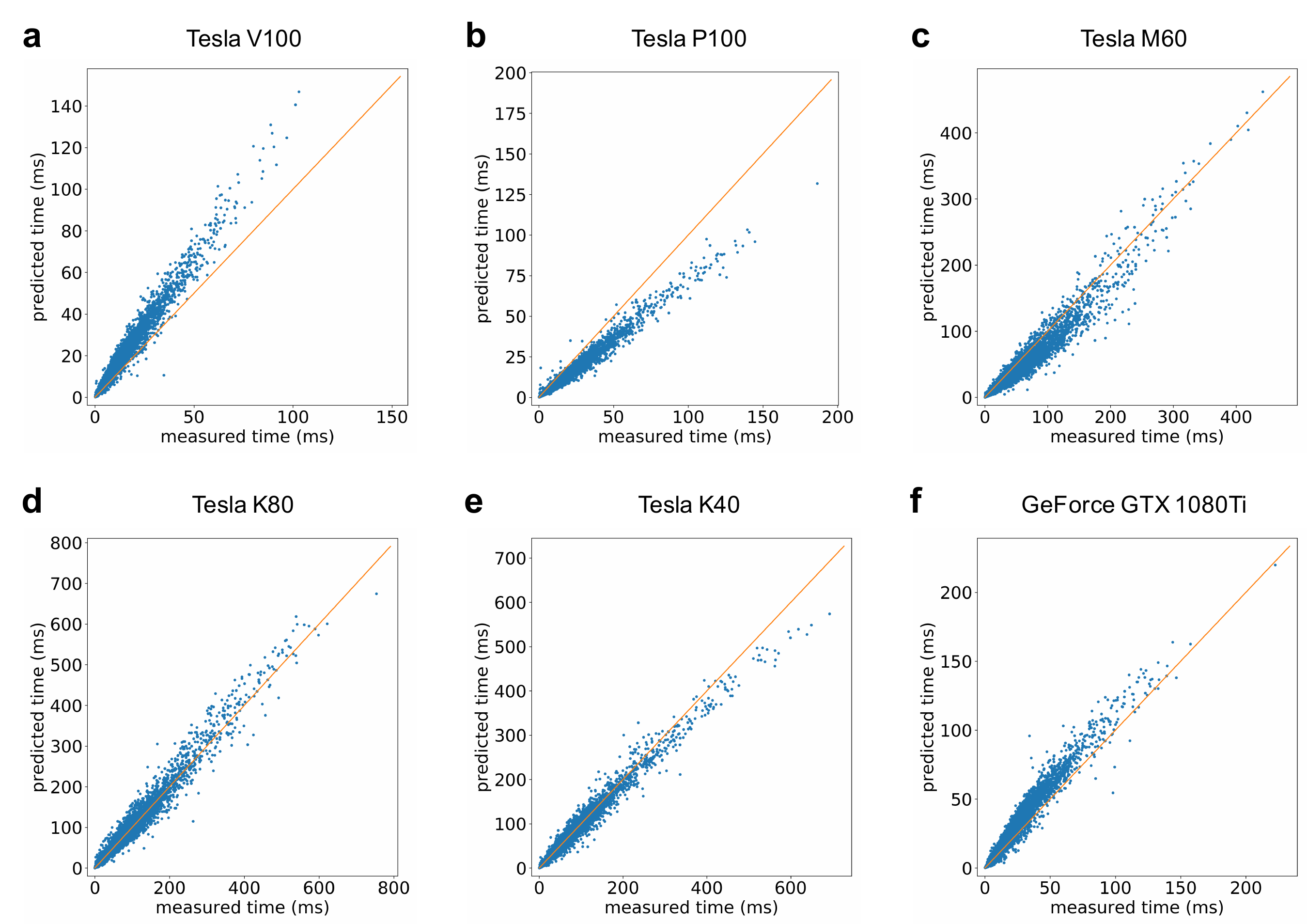}}
	\caption{Predicted time for convolutions plotted against measured time for a model that has been trained on data from a set of GPUs excluding the one tested. \textbf{a)} Predictions of NVIDIA Tesla V100 performance from data of other GPUs. The RMSE is 3.65 ms. \textbf{b)} Predictions of NVIDIA Tesla P100 performance from data of other GPUs. The RMSE is 4.32 ms. \textbf{c)} Predictions of NVIDIA Tesla M60 performance from data of other GPUs. The RMSE is 8.09 ms. \textbf{d)} Predictions of NVIDIA Tesla K80 performance from data of other GPUs. The RMSE is 7.94 ms. \textbf{e)} Predictions of NVIDIA Tesla K40 performance from data of other GPUs. The RMSE is 7.00 ms. \textbf{f)} Predictions of NVIDIA Geforce GTX1080Ti performance from data of other GPUs. The RMSE is 4.24 ms.}
	\label{fig:mixedModel}
\end{figure*}

In Figure~\ref{fig:mixedModel} we evaluate the ability for this mixed model to predict the execution times for training on an unseen card. We assessed the prediction accuracy for every GPU separately, i.e. for each of the six GPUs we trained a model using data from the other five GPUs. While we observe some small but systematic discrepancies between predicted and measured results, the overall results are very encouraging, especially since additional training data from other GPUs can be expected to further improve the accuracy. The uncertainty of predictions can be expected to be particularly large for GPUs with specifications that don't fall into the range of known hardware. Therefore, we observe an over-prediction of the execution time for the NVIDIA Tesla V100, the newest GPU that has been tested here. The same problem does not arise at the lower border of the performance spectrum, since the two oldest GPUs, the NVIDIA Tesla  K80 and K40, have quite similar characteristics.

\subsection{Inferring a full model prediction}
\begin{figure*}[htbp]
	\centerline{\includegraphics[width=1.0\textwidth]{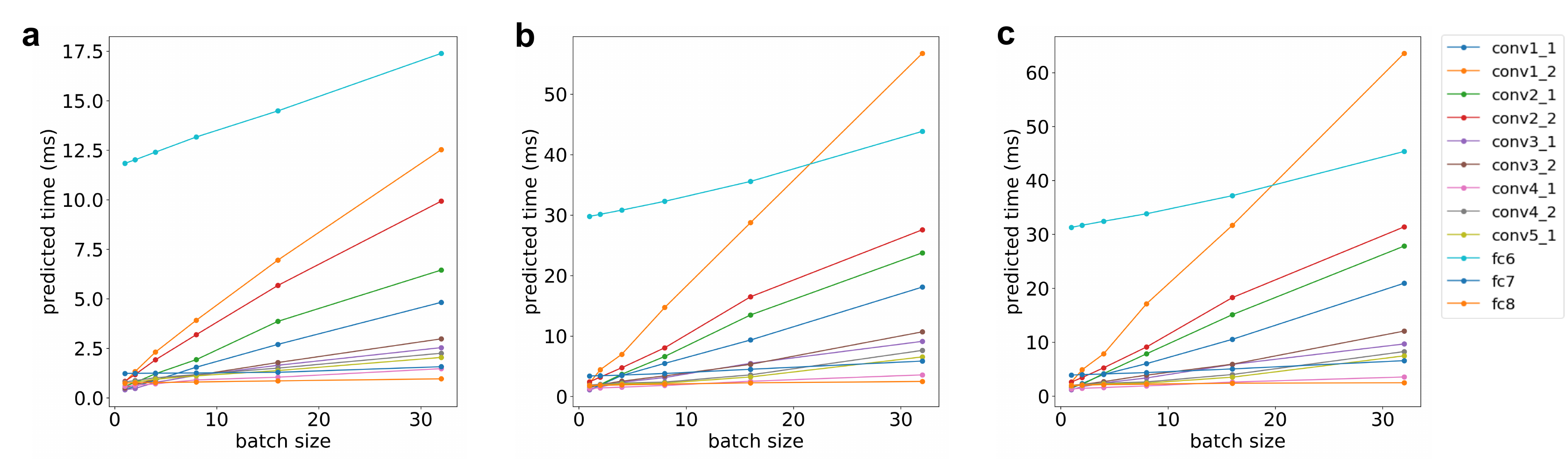}}
	\caption{Predicted time for different layers of the VGG-16 model on a NVIDIA Tesla V100 GPU. \textbf{a)} Only forward pass \textbf{b)} Forward and backward pass with standard stochastic gradient descent. \textbf{c)} Forward and backward pass with Adam optimiser.}
	\label{fig:layerVGG}
\end{figure*}

Here we aim to predict the execution time for an entire epoch of a deep learning network. We choose the VGG-16 \cite{Simonyan2014} network to test our prediction model. We first predict the execution time for a forward or a forward and backward pass through each layer within the network. Then, we combine these results to predict the execution time for the entire deep neural network. Our model is trained with the data from all six different GPUs that we have been using. 

It should be noted that although here we evaluate our model against benchmarks of the VGG-16 model, our approach is as equally valid for predicting the execution time for any deep learning network that primarily depends on convolutions and matrix multiplications. Similar to Section~\ref{general_model} this methodology can also be expanded to different hardware setups.

Our model allows for inferring the execution time for each layer of a deep learning model. It therefore can be used to predict the effect of arbitrary changes to the model architecture or parameters such as the batch size or the optimiser used (see Figure~\ref{fig:layerVGG} where we vary the batch size used and the optimizer). 

Our deep learning predictor approach nicely demonstrates the different scaling properties of the compute time for the different batch sizes and different layers. While the batch size only has a small effect on the execution time for fully connected layers, the execution time for convolutional layers grows strongly with the batch size. Therefore, fully connected layers dominate the overall processing time in the case of small batch sizes, while convolutional layers dominate at larger batch sizes. These different characteristics can be explained by the large amount of weights that need to be written to GPU memory for fully connected layers, while the execution time for convolutional layers is dominated by the computation time.

\begin{figure}[tbp]
	\centerline{\includegraphics[width=0.45\textwidth]{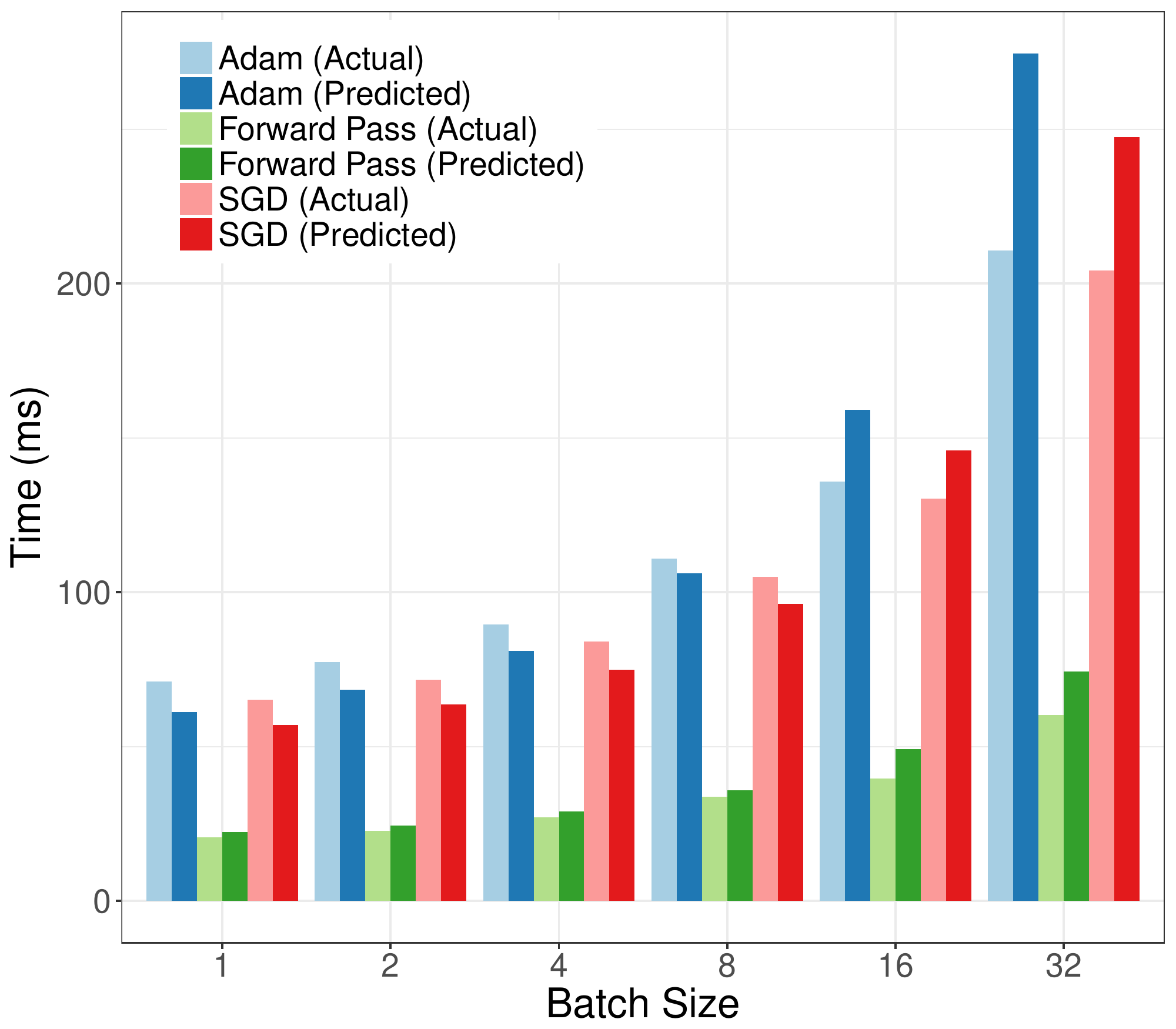}}
	\caption{Predicted versus actual execution time for VGG-16.}
	\label{fig:graph}
\end{figure}

We can now combine individual layer times to predict execution times for forward and backward passes of independent batches through a complete deep neural network. Figure~\ref{fig:graph} depicts the comparison between predicted and actual execution time for a range of batch sizes, using forward pass only, SGD and Adam optimizer. While these results show some variation especially for large batch sizes, they allow us to draw conclusions on the computational complexity of the given model and estimate the training time for one epoch. Moreover, it allows us to predict the effects that different hardware and different model characteristics, such as the optimizer, have on the resulting execution time. 

\section{Conclusions and future directions}
\label{conc}

In this work we have demonstrated a deep learning model that is capable of accurately predicting the required execution time for a wide range of the most frequently used components of neural networks. This model extends upon previous works by additionally considering nonlinear components of neural networks, which up to this point have been largely ignored. The results from this model can be used to predict execution times for complete deep neural networks. Thereby, the model can provide a good foundation for informed choices when selecting appropriate hardware to train a model or infer predictions, while additionally helping to inform decisions around model design and layout.

Our model provides an extensible framework, allowing anyone to easily incorporate their own model architecture along with hardware specific training data. It can also be easily extended to different machine learning frameworks and programming languages. Furthermore, with additional data this model can be expanded to 16 bit floating point as well as lower precision fixed point computations. This allows for straightforward generation of prediction models either for specific use cases or, where the data is available, a more general model that is capable of predicting over a broader range of different configurations.

In future work, our model will benefit from additional benchmarking on a wider array of hardware along with further investigation into an expanded set of appropriate `features' used for training the model. This additional work will eventually provide a highly flexible model, able to generalise very well across varying network models, data sizes and hardware configurations. We aim to make this model framework a tool that can be used by researchers as well as interested individuals and industry. We will provide all code relevant and make it available on an open platform, allowing the sharing of benchmark results to further improve generalization of the model.


\section*{Acknowledgment}
We would like to thank Innovate UK and the European Regional Development Fund for their contribution towards making this work possible.


\bibliographystyle{IEEEtran}
\bibliography{./ML.bib}

\begin{thebibliography}{10}
\providecommand{\url}[1]{#1}
\csname url@samestyle\endcsname
\providecommand{\newblock}{\relax}
\providecommand{\bibinfo}[2]{#2}
\providecommand{\BIBentrySTDinterwordspacing}{\spaceskip=0pt\relax}
\providecommand{\BIBentryALTinterwordstretchfactor}{4}
\providecommand{\BIBentryALTinterwordspacing}{\spaceskip=\fontdimen2\font plus
\BIBentryALTinterwordstretchfactor\fontdimen3\font minus
  \fontdimen4\font\relax}
\providecommand{\BIBforeignlanguage}[2]{{%
\expandafter\ifx\csname l@#1\endcsname\relax
\typeout{** WARNING: IEEEtran.bst: No hyphenation pattern has been}%
\typeout{** loaded for the language `#1'. Using the pattern for}%
\typeout{** the default language instead.}%
\else
\language=\csname l@#1\endcsname
\fi
#2}}
\providecommand{\BIBdecl}{\relax}
\BIBdecl

\bibitem{hu2017}
J.~Hu, L.~Shen, and G.~Sun, ``Squeeze-and-excitation networks,'' \emph{arXiv
  preprint arXiv:1709.01507}, vol.~7, 2017.

\bibitem{silver}
D.~Silver, A.~Huang, C.~J. Maddison, A.~Guez, L.~Sifre, G.~Van Den~Driessche,
  J.~Schrittwieser, I.~Antonoglou, V.~Panneershelvam, M.~Lanctot \emph{et~al.},
  ``Mastering the game of go with deep neural networks and tree search,''
  \emph{Nature}, vol. 529, no. 7587, pp. 484--489, 2016.

\bibitem{Qi2016}
\BIBentryALTinterwordspacing
H.~Qi, E.~R. Sparks, and A.~Talwalkar, ``Paleo: A performance model for deep
  neural networks,'' 2016. [Online]. Available:
  \url{https://openreview.net/pdf?id=SyVVJ85lg}
\BIBentrySTDinterwordspacing

\bibitem{Krizhevsky2012}
\BIBentryALTinterwordspacing
A.~Krizhevsky, I.~Sutskever, and G.~E. Hinton, ``Imagenet classification with
  deep convolutional neural networks,'' in \emph{Advances in neural information
  processing systems}, 2012, pp. 1097--1105. [Online]. Available:
  \url{https://papers.nips.cc/paper/4824-imagenet-classification-with-deep-convolutional-neural-networks.pdf}
\BIBentrySTDinterwordspacing

\bibitem{He2015}
\BIBentryALTinterwordspacing
K.~He, X.~Zhang, S.~Ren, and J.~Sun, ``Deep residual learning for image
  recognition,'' 2015. [Online]. Available:
  \url{https://arxiv.org/pdf/1512.03385.pdf}
\BIBentrySTDinterwordspacing

\bibitem{Zhang2018}
\BIBentryALTinterwordspacing
Q.~Zhang, L.~T. Yang, Z.~Chen, and P.~Li, ``A survey on deep learning for big
  data,'' \emph{Information Fusion}, vol.~42, pp. 146 -- 157, 2018. [Online].
  Available:
  \url{http://www.sciencedirect.com/science/article/pii/S1566253517305328}
\BIBentrySTDinterwordspacing

\bibitem{Coleman2017}
\BIBentryALTinterwordspacing
C.~Coleman, D.~Narayanan, D.~Kang, T.~Zhao, J.~Zhang, L.~Nardi, P.~Bailis,
  K.~Olukotun, C.~R{\'e}, and M.~Zaharia, ``Dawnbench: An end-to-end deep
  learning benchmark and competition,'' \emph{Training}, vol. 100, no. 101, p.
  102, 2017. [Online]. Available:
  \url{http://dawn.cs.stanford.edu/benchmark/papers/nips17-dawnbench.pdf}
\BIBentrySTDinterwordspacing

\bibitem{Adolf2016}
\BIBentryALTinterwordspacing
R.~Adolf, S.~Rama, B.~Reagen, G.-Y. Wei, and D.~Brooks, ``Fathom: Reference
  workloads for modern deep learning methods,'' in \emph{Workload
  Characterization (IISWC), 2016 IEEE International Symposium on}.\hskip 1em
  plus 0.5em minus 0.4em\relax IEEE, 2016, pp. 1--10. [Online]. Available:
  \url{https://arxiv.org/pdf/1608.06581.pdf}
\BIBentrySTDinterwordspacing

\bibitem{jouppi2017}
N.~P. Jouppi, C.~Young, N.~Patil, D.~Patterson, G.~Agrawal, R.~Bajwa, S.~Bates,
  S.~Bhatia, N.~Boden, A.~Borchers \emph{et~al.}, ``In-datacenter performance
  analysis of a tensor processing unit,'' in \emph{Computer Architecture
  (ISCA), 2017 ACM/IEEE 44th Annual International Symposium on}.\hskip 1em plus
  0.5em minus 0.4em\relax IEEE, 2017, pp. 1--12.

\bibitem{relu}
V.~Nair and G.~E.~Hinton, ``Rectified linear units improve restricted boltzmann
  machines vinod nair,'' in \emph{Proceedings of ICML}, vol.~27, 06 2010, pp.
  807--814.

\bibitem{l2r}
A.~Neumaier, ``Solving ill-conditioned and singular linear systems: A tutorial
  on regularization,'' vol.~40, 01 1998.

\bibitem{adam}
D.~P. Kingma and J.~L. Ba, ``Adam: Amethod for stochastic optimization,'' in
  \emph{Proceedings of the 3rd International Conference on Learning
  Representations (ICLR)}, 2015.

\bibitem{linear}
F.~Galton, ``Regression towards mediocrity in hereditary stature,''
  \emph{Journal of the Anthropological Institute}, vol.~15, pp. 246--263, 1886.

\bibitem{kiefer1952}
\BIBentryALTinterwordspacing
J.~Kiefer and J.~Wolfowitz, ``Stochastic estimation of the maximum of a
  regression function,'' \emph{Ann. Math. Statist.}, vol.~23, no.~3, pp.
  462--466, 09 1952. [Online]. Available:
  \url{https://doi.org/10.1214/aoms/1177729392}
\BIBentrySTDinterwordspacing

\bibitem{Simonyan2014}
\BIBentryALTinterwordspacing
K.~Simonyan and A.~Zisserman, ``Very deep convolutional networks for
  large-scale image recognition,'' \emph{arXiv preprint arXiv:1409.1556}, 2014.
  [Online]. Available: \url{https://arxiv.org/pdf/1409.1556.pdf}
\BIBentrySTDinterwordspacing

\end{thebibliography}

\end{document}